\documentclass[letterpaper]{article}
\usepackage[preprint]{amlc_2023}
\usepackage[page]{appendix} 
\usepackage{multirow}

\usepackage{times}

\usepackage{mathtools}
\usepackage{graphicx}
\usepackage{mathrsfs}
\usepackage{amsmath}
\usepackage{amssymb}
\usepackage{natbib}
\usepackage{soul,color}
\usepackage{algorithm,algcompatible}
\usepackage{tabularx}
\usepackage{mathrsfs}
\usepackage{multirow}
\usepackage{bbm}
\usepackage{amsthm}
\usepackage{bbold}
\usepackage{color,soul}

\title{Teacher-Student Learning on Complexity in Intelligent Routing}

\author{ {\bf Shu-Ting~Pi} \\
Amazon   \\
Cupertino, CA 95014\\
shutingp@amazon.com\\
\And
{\bf Michael Yang}   \\
Amazon \\
Seattle, WA 98109    \\
abyang@amazon.com
\And
{\bf Yuying Zhu}   \\
Amazon \\
Seattle, WA 98109    \\
abyang@amazon.com
\And
{\bf Qun Liu}  \\
Amazon           \\
Seattle, WA 98109 \\
qunlin@amazon.com
}

\begin{document}

\maketitle
\begin{abstract}
Customer service is often the most time-consuming aspect for e-commerce websites, with each contact typically taking 10-15 minutes. Effectively routing customers to appropriate agents without transfers is therefore crucial for e-commerce success. To this end, we have developed a machine learning framework that predicts the complexity of customer contacts and routes them to appropriate agents accordingly. The framework consists of two parts. First, we train a teacher model to score the complexity of a contact based on the post-contact transcripts. Then, we use the teacher model as a data annotator to provide labels to train a student model that predicts the complexity based on pre-contact data only. Our experiments show that such a framework is successful and can significantly improve customer experience. We also propose a useful metric called "complexity AUC" that evaluates the effectiveness of customer service at a statistical level.

\end{abstract}

\section{Introduction}
 E-commerce businesses face a significant influx of customer service requests on a daily basis, making it one of the most time-consuming aspects of the industry. Processing each request typically takes 10-15 minutes. To illustrate, consider a company with 1 million users, where even a small percentage of customers requiring assistance, such as 0.1\%, would result in over 10000 minutes spent on servicing customers daily. For companies with hundreds of millions of users, this number becomes overwhelmingly high and can become a bottleneck to expanding their business. Effectively guiding customers to the right service agents is crucial for achieving success in the industry, as emphasized by \cite{universal_model}. However, the support requirements of customers can vary significantly depending on the nature of their issues. Complex problems may demand the expertise of senior agents, while junior agents can effectively handle simpler issues such as returns or refunds. Misdirecting a customer with a complex issue to a junior agent can result in poor customer experiences, involving transfers or repeated contacts. On the other hand, directing a customer with a simple issue to a senior agent is costly for the business and restricts the availability of senior agents to address complex problems. Consequently, there is a growing demand for machine learning tools that can identify the complexity of customer requests from the beginning and appropriately route them to the designated agent, thereby enhancing the overall customer experience.

However, defining "complexity" is challenging, given its highly subjective nature. What may be considered difficult for a junior agent may not pose a challenge for a senior agent. Although consensus-based data annotation by senior agents could provide reliable and professional labels, the process of labeling thousands, if not millions, of contacts to create a dataset is slow and incredibly expensive.

We propose a unique teacher-student learning framework to tackle the issue at hand, which capitalizes on complexity to establish an efficient routing system. Our method involves training a teacher model using transcripts of conversations between customers and agents to assess the complexity of each contact. Since post-contact data cannot be used directly for routing purposes, we use it as a labeling mechanism to train a student model that can predict complexity based on pre-contact features such as the customer's profile and purchase history. Our experiments demonstrate that this framework can significantly enhance customer experience across various service metrics by directing high complexity contacts to senior agents and low complexity contacts to junior agents.

Additionally, we present a novel metric called "Complexity AUC," which is developed by linking the distribution of complexity scores with a benchmark distribution and can accurately measure the routing system's effectiveness at a statistical level.

The article's structure is outlined below: In Section 2, we elaborate on the development of the teacher model. In Section 3, we explain how the teacher model can be utilized as a data annotator to train a student model to predict the contact's complexity from the outset. In Section 4, we introduce the concept of Complexity AUC and provide demonstrations. Finally, we draw conclusions in Section 5.

\textbf{Related Works} 
Our work draws significant parallels with previous studies that employ information theory quantities to characterize input data properties. The information bottleneck concept, for instance, implements similar approaches and calculates mutual information between intermediate and final layer activations in a neural network to illustrate how data point noise is filtered through the network \citep{bottleneck1,bottleneck2}. While our method is similar, we use KL divergence and ensemble boosting.

Another area closely related to our project is the classification of text difficulty, which examines the readability of articles for different age groups of children to aid educators in crafting appropriate materials. Our model is a prime example of how text difficulty can be linked to customer service. Other related studies on machine learning applications in this field can be found in \citep{difficulty1} and \citep{difficulty2}. Additionally, the implementation of text difficulty in business settings such as healthcare is a noteworthy topic \citep{difficulty3}.

\begin{figure*}
  \centering
  \includegraphics[width=1.0\textwidth]{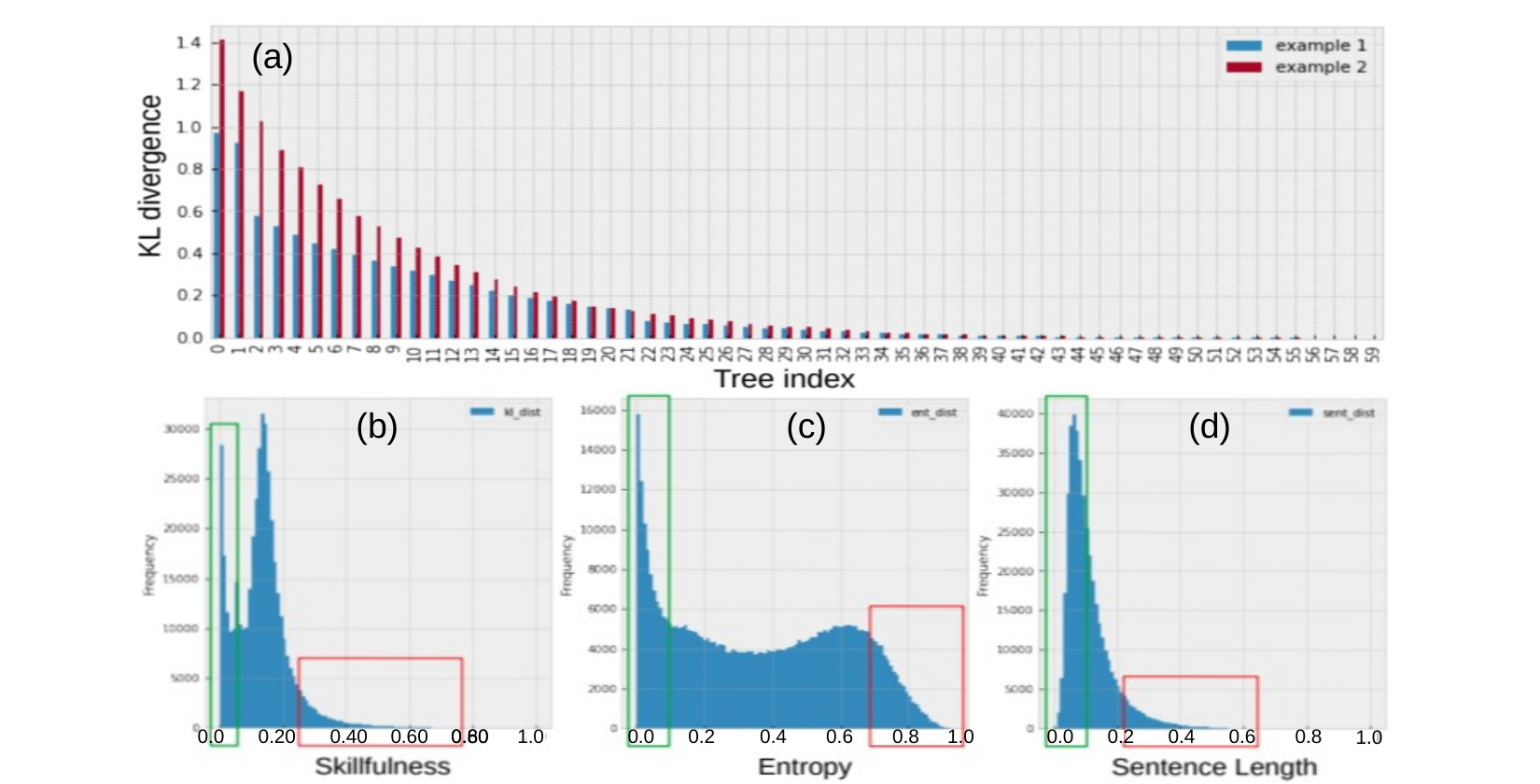}
  \caption{(a) We demonstrate the KL divergence boosting function with 60 trees for two different examples. The function decays faster for example 1 than for example 2, resulting in a smaller integral or skillfulness. (b)-(d) The skillfulness, entropy, and sentence length (all normalized to 1) distributions of 500K contacts are shown. Higher complexity contacts are represented by the red regions, while lower complexity contacts are represented by the green regions.}\label{boosting_function}
\end{figure*}

\section{Teacher Model}
Our objective is to develop a machine learning model that can evaluate the complexity of a contact based on the conversation transcript between the customer and service agent. We will demonstrate that complexity can be characterized using certain quantities derived from a machine learning model equipped with knowledge of the relevant business.

\subsection{Hypothesis of Complexity}
Our approach to defining complexity involves initially training a machine learning classifier that employs the contact transcripts as inputs to predict the corresponding product or service labels (see Appendix for an example transcript). We presume that this classifier possesses the fundamental domain knowledge of the core business. If this is true, we conjecture that complexity can be described using the following quantities:
\begin{itemize}
    \item \textbf{Length}: Contacts that are high in complexity generally necessitate multiple steps to diagnose the underlying issues, resulting in longer transcripts than those with lower complexity. This can be mathematically represented by the number of sentences spoken by the agent(s) involved in the contact, as follows:
    \[\mathcal{L} = \sum_{i=1}^{N}\delta_{S_{i}, agent}\]
    Here, $\delta$ denotes the Kronecker delta function, $S_{i}$ represents the speaker of the $i$-th sentence in the contact transcript, and $N$ denotes the total number of sentences in the transcript. The value of $\mathcal{L}$ represents the total number of sentences spoken by the agent(s) in the contact, which can be used to measure the contact's complexity.
    
    \item \textbf{Uncertainty}: High-complexity contacts can be more challenging to comprehend, which can result in a greater level of uncertainty for the classifier regarding the customers' issues. This can be mathematically represented by the entropy of the classifier's output distribution, as follows\citep{info_entropy}:
    \[\mathcal{H} = -\sum_{c}p_{c}log(p_{c}) \]
    Here, $p(c)$ denotes the probability of a particular class $c$ being predicted by the classifier, and $H$ represents the entropy of the classifier's output distribution. A higher entropy value indicates greater uncertainty in the classifier's predictions, which may occur with high-complexity contacts.
    
    \item \textbf{Skillfulness}: Contacts that are high in complexity generally require a higher degree of skill to resolve, necessitating a highly proficient model that can effectively handle such contacts. This can be mathematically represented by the integral of the KL divergence\citep{kl_divergence} between the outputs of using fewer trees and more trees in a gradient boosting trees model\citep{LightGBM}, as follows:
    \[\mathcal{S}=\sum_{i}^{M}\phi(i)=\sum_{i}^{M}D_{KL}(P_{i}||P_{M}) \]
    Here, $P_{i}$ denotes the output distribution using the first $i$ trees, and $M$ represents the total number of trees in the model. It is essential to note that the KL boosting function $\phi(i)$ exhibits a slow decay to zero when more knowledge or skills are necessary, and a rapid decay to zero when less knowledge or skills are required. Fig.1(a) demonstrates two examples of how the divergence boosting function behaves with more (red) and less (blue) skills.
\end{itemize}

\subsection{The Dataset}
Until now, there has been a dearth of publicly available datasets pertaining to customer service transcripts. To address this gap, we collaborated with customer service team to launch this research project, leveraging a dataset from an online chat system that allows customers to communicate with customer service representatives. We collected a total of 500,000 contact transcripts during 2022, each contact is associated with a unique label indicating the product or service discussed by the customer and agent. It is essential to note that the dataset only comprises customer text data, with all sensitive information, such as names and account details, anonymized to preserve confidentiality before sharing with researchers. Despite the dataset being pulled from a specific database, the methodology introduced in this article can be adapted to other scenarios as well.

\begin{figure*}
  \centering
  \includegraphics[width=1.0\textwidth]{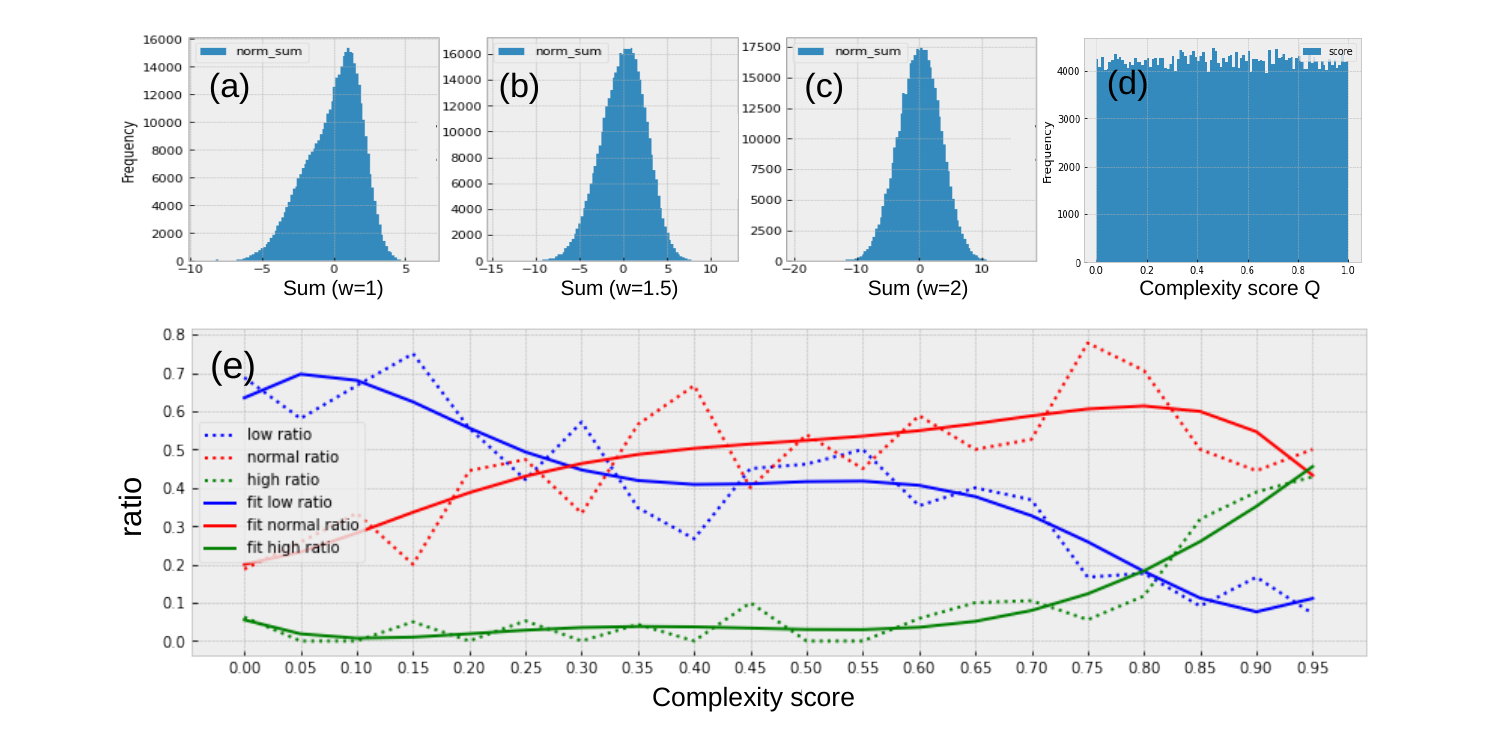}
  \caption{((a)-(c) We present the distribution of the sum of the three complexity measures using various weights on length.
(d) We convert the distribution from (c) into a uniform distribution using quantile transformation.
(e) The probability of finding high/medium/low complexity in different scores is shown. Over 400 ground truth labels were generated via senior agents. The dash lines represent true numbers, while the solid lines are fitted curves using polynomial functions. The blue line represents low complexity, the red line represents medium complexity, and the green line represents high complexity. }\label{flow_chart}
\end{figure*}

\subsection{Complexity Score} 
We employed a LightGBM\citep{LightGBM} classifier with 60 boosting trees to predict the label of a product or service among 152 classes using TF-IDF\citep{tf-idf} vectors generated from 500K transcripts. The resulting $\mathcal{L}$, $\mathcal{H}$ and $\mathcal{S}$ distributions are illustrated in Fig.1(b)-(d), which demonstrate a long-tailed structure with high-complexity contacts located in the tail (red) region and low-complexity contacts in the head (green) region. However, using three different complexity measures can be complicated for businesses. Therefore, we aimed to introduce a single variable that represents complexity in a more straightforward manner. To achieve this, we converted all attributes to a normal distribution through quantile transformation\citep{quantile}, ensuring that they have the same scale and can be combined safely. We denoted the transformed variables as $\mathcal{L}^{N}$, $\mathcal{H}^{N}$, and $\mathcal{S}^{N}$, and we calculated a single variable $\mathcal{Q}$ using the following formula:
\[\mathcal{Q} = \mathcal{T^U_G}(w\times\mathcal{L^N+H^N+S^N})\]
Here, we introduced a weight number $w$ on $\mathcal{L}^{N}$ for two reasons: first, short transcripts are unlikely to be high-complexity contacts, and giving a higher weight to this attribute can increase the model's precision. Second, we found that an appropriate weight on $\mathcal{L}^{N}$ can transform the distribution of the sum in the bracket from a skewed distribution to a Gaussian distribution that satisfies our expectation, where only a few contacts are extremely simple or extremely complex, as shown in Fig.2(a)-(c). We determined that using $w=2$ resulted in a distribution closest to a Gaussian distribution and will use this value unless otherwise stated. Additionally, we utilized another quantile transformation, denoted as $\mathcal{T^U_G}$, to convert the Gaussian-like distribution to a uniform distribution within the bracket, enabling us to have a complexity score between 0 to 1 as shown in Fig.2(d). The complexity score $\mathcal{Q}$ represents the percentile of a contact's complexity among all contacts. For example, a complexity score of $\mathcal{Q}=0.95$ indicates that the contact's complexity is higher than that of 95\% of all contacts.

\subsection{Validation}
We validated our complexity score using both objective and subjective metrics. For objective metrics, we visually inspected 400 examples of high- and low-complexity contacts with complexity scores $\mathcal{Q}>0.95$ and $\mathcal{Q}<0.05$, respectively. Our survey showed that only 26\% of high-complexity contacts were resolved, while 62\% required at least one transfer. In contrast, 85\% of low-complexity contacts were resolved, and only 11\% required at least one transfer. This significant difference between the two groups demonstrates the effectiveness of our complexity score.

For subjective metrics, we collaborated with senior service agents to label the complexity of 400 examples. We ensured at least two agents reviewed each contact and reached a consensus label of "low," "normal," or "high" complexity. Fig.2(e) presents the results of binning the complexity scores into 20 intervals and calculating the probability of finding a low, normal, or high complexity example in each interval. The likelihood of finding a low complexity contact gradually decreases to 0.1 as the complexity score increases, while the probability of finding a high complexity contact rapidly increases to 0.5 after a complexity score of 0.6. This confirms the consistency of our complexity score with the agents' understanding of contact complexity.

\section{Student Model}
\subsection{Model Training}

We developed a teacher model that analyzes customer-agent conversation transcripts to calculate their complexity. However, as these transcripts are only available after each contact is finished, i.e. a post-contact feature, they cannot be used for intelligent routing. To solve this issue, we used the complexity scores generated by our teacher model as ground truth labels and trained a machine learning model using pre-contact data only. Our proof of concept experiment aims to route high complexity contacts to senior agents, as these interactions have the greatest impact on customer experience.

To train the model, we followed these steps:
\begin{itemize}
\item First, we defined high complexity contacts as those with a complexity score $\mathcal{Q}$ of 0.8 or higher and labeled them as $y=1$. Contacts with a score below 0.8 were labeled as $y=0$. We then trained a binary classifier based on these labels.
\item We gathered more than 100 pre-contact features, such as customer profiles, purchase history, contact history, and device usage, all of which were accessible at the beginning of each interaction. These features were employed to train a gradient boosting tree model\citep{LightGBM,xgboost,catboost}. To handle the categorical features, we used the entity embedding method proposed by \citep{enity_embedding} to convert them into continuous variables.
\item Using this setup, we found that our model achieved a precision of 0.56 and recall of 0.28. In comparison, a model without entity embedding achieved a precision of 0.54 and recall of 0.05. Our results demonstrate that using entity embedding can significantly improve the recall rate and overall performance of the model.
\end{itemize}

\subsection{Experimental Results}
We worked together with customer service team to assess the performance of our model in real-world scenarios. Our approach involved randomly selecting 4000 contacts that were identified as having high complexity by the student model. We divided the contacts into two groups, with 50\% of them being directed to normal agents as a control group and the other 50\% to senior agents as the treatment group.

\subsubsection{Distributions of Complexity Score}
We computed complexity scores using the teacher model for various groups to gain insights into our model's real-world business performance. Complexity is a metric of how agents respond to a contact, not an inherent attribute of the contact. Figure 3(a)-(c) displays the teacher model's complexity score distributions for different groups, including (a) the background group, consisting of 10K randomly selected contacts (b) the control group, and (c) the treatment group.

The teacher model assumes a uniform distribution of overall complexity scores, so the complexity score distribution for (a) is expected to be uniform. The distribution of complexity scores for (b) is negatively skewed, indicating that the student model correctly captures high complexity contacts. This suggests that contacts in the control group were lengthier, more uncertain, and required more skills from regular agents than contacts in (a). In (c), the distribution of complexity scores tends to be much lower than (b). Senior agents have more experience dealing with high complexity contacts, allowing them to handle such contacts more efficiently, with less ambiguity and fewer required skills. This reduces the complexity of the conversations, enabling the classifier to make the right decision with fewer trees and higher confidence. As a result, the complexity score in the treatment group is much lower than that in the control group. In summary, our model can filter out high complexity contacts, and routing them to senior agents can significantly improve the overall customer experience, as seen in the difference between (b) and (c).

\subsubsection{Routing Metrics}
Upon analyzing the advantages of directing high complexity contacts to senior agents, we conducted a comparison of essential metrics. Remarkably, the treatment group displayed a transfer rate approximately 53\% lower than the control group. In terms of multi-transfers, the treatment group experienced a reduction of over 95\% compared to the control group, effectively eliminating the majority of multi-transfers. Additionally, the treatment group showcased an average handle time that was 13\% lower than the control group. These findings validate the effectiveness of our teacher-student model, showcasing its capacity to significantly enhance the customer experience by reducing transfers and decreasing average handle times.

\section{Complexity AUC}
\subsection{Dual Transformation}

\begin{figure*}
  \centering
  \includegraphics[width=1.0\textwidth]{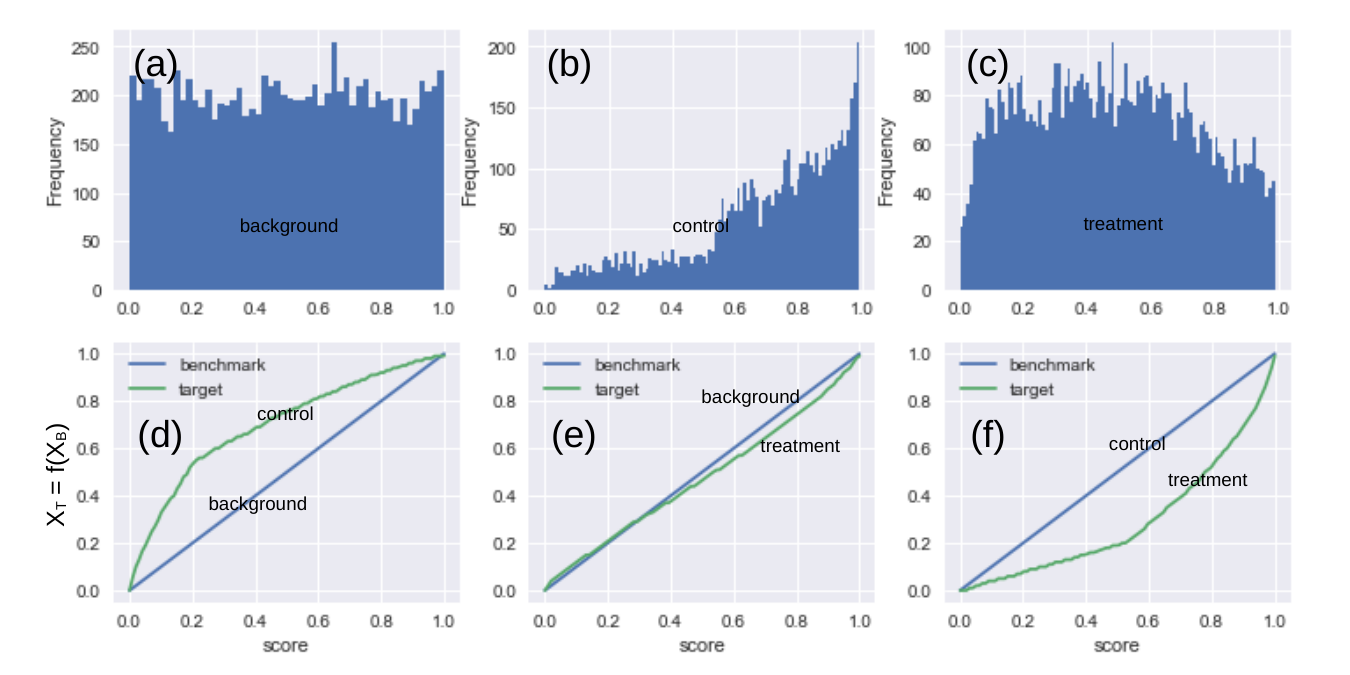}
  \caption{(a)-(c) We show the distribution of complexity scores for the background group, control group, and treatment group, respectively.
(d)-(f) The dual transformation curve is displayed for the (background, control), (background, treatment), and (control, treatment) pairs used as the benchmark and target groups. }\label{auc_curve}
\end{figure*}

In Section 3, we presented the variations in complexity score distributions among different groups. This statistical property can be utilized to establish a novel metric, known as "Complexity AUC" (area under the curve), which is analogous to the traditional AUC\citep{auc-1,auc-2,auc-3,AUC_mu}, for assessing the routing business's effectiveness. To achieve this, we must introduce the notion of dual transformation.

Suppose we have two groups of contacts, the target group and the benchmark group, and we want to compare whether the service agents handle the target group more effectively than the benchmark group. The first step in defining Complexity AUC is to find a non-linear transformation that maps the complexity score distribution of the benchmark group to that of the target group. In other words, we need to find a transformation that makes the benchmark distribution identical to the target distribution. 

However, directly searching for such a transformation is difficult. One possible solution is to first map both distributions to a normal distribution. This is achieved by using operators $\mathcal{T}_{B}^{N}$ and $\mathcal{T}_{T}^{N}$ to transform $\mathcal{Q}_{B}$ and $\mathcal{Q}_{T}$, respectively, to a normal distribution with mean = 0 and variance = 1, denoted as $\mathcal{N}(\mu=0,\sigma=1)$:
\begin{align*}
\mathcal{T}_{B}^{N}\mathcal{Q}_{B} = \mathcal{N}(\mu=0,\sigma=1)\\
\mathcal{T}_{T}^{N}\mathcal{Q}_{T} = \mathcal{N}(\mu=0,\sigma=1)
\end{align*}

Here, $\mathcal{Q}_{B}$ and $\mathcal{Q}_{T}$ are the benchmark and target distributions, respectively. If the inverse operator $\mathcal{T}_{T}^{N^{-1}}$ exists, we can define a dual transformation $\mathcal{T}_{B}^{T}$ that transforms $\mathcal{Q}_{B}$ to $\mathcal{Q}_{T}$:
\[
    \mathcal{T}_{B}^{T} \mathcal{Q}_{B} = \mathcal{T}_{T}^{N^{-1}}\mathcal{T}_{B}^{N}\mathcal{Q}_{B} = \mathcal{Q}_{T}
\]
Here, $\mathcal{T}_{B}^{N}$ is an operator that transforms $\mathcal{Q}_{B}$ to $\mathcal{N}$, and $\mathcal{T}_{T}^{N}$ is an operator that transforms $\mathcal{Q}_{T}$ to $\mathcal{N}$. 

In general, $\mathcal{T}_{T}^{N^{-1}}$ always exists provided that $\mathcal{Q}_{T}$ is differentiable, which can be achieved through various methods such as Power transformation \citep{power-1, power-2} or Quantile transformation \citep{quantile}. Therefore, we can always find the dual transformation $\mathcal{T}_{B}^{T}$ without difficulty. The meaning of $\mathcal{T}_{B}^{T}$ is also clear. It tells us how to shift the complexity scores in the benchmark group to make its distribution identical to that of the target group.

\subsection{Area Under Curve}
Let's explore the mathematical properties of $\mathcal{T}_{B}^{T}$. Since $\mathcal{T}_{B}^{T}$ is essentially a function that transforms a complexity score in the benchmark group to the corresponding score in the target group, $x_{T} = f(x_{B}) \equiv \mathcal{T}_{B}^{T}x_{B}$, where $x_{B}$ is a complexity score in the benchmark group. If we change $x_{B}$ linearly from 0 to 1, we obtain a curve in a 2D plane where the x-axis represents $x_{B}$ and the y-axis represents $f(x_{B})$.
   
When the target distribution is identical to the benchmark distribution, the operator $\mathcal{T}_{T}^{B}$ is an identity operator, resulting in a straight line connecting (0,0) and (1,1) with an area under the curve of 0.5, indicating that the complexity of the target group is statistically similar to that of the benchmark group. However, a concave curve for $f(x_{B})$ results in an area greater than 0.5, indicating more negative skewness and greater complexity in the target group, while a convex curve leads to an area less than 0.5, indicating more positive skewness and less complexity. When contact complexity is lower than the background complexity, agents can handle the contact more efficiently, confidently, and with less ambiguity, reducing the likelihood of transfers or unresolved contacts. Therefore, the area under the curve (Complexity AUC) evaluates how effectively agents manage specific contact groups. The following guidelines may be used to interpret Complexity AUC:

\begin{equation*}
\text{Complexity AUC } : \begin{cases}
> 0.5 & \text{$\mathcal{G}_t$ more effective than $\mathcal{G}_b$}\\
= 0.5 & \text{$\mathcal{G}_t$ equally effective compared to $\mathcal{G}_b$}\\
< 0.5 & \text{$\mathcal{G}_t$ less effective than $\mathcal{G}_b$}
\end{cases}        
\end{equation*}
, where $\mathcal{G}_t$ and $\mathcal{G}_b$ refers to target group and benchmark group respectively.

\subsection{Experimental Results}
Figure 3(d)-(f) depicts the dual transformation curve using different benchmark and target groups. Each subfigure contains two curves. The blue curve represents the case where the benchmark and target distributions are identical, resulting in an identity transform and a straight line as a reference. On the other hand, the green curve uses the true distribution of the target group, resulting in deviation from the blue curve. In Figure 3(d), the background group is used as the benchmark and the control group as the target. As shown in Fig.3(b), the control group has the highest complexity contacts for average agents, hence the green curve is concave with an AUC of 0.69. We define the complexity effectiveness as $\epsilon=1-AUC/0.5$, and the effectiveness of the control group is -39.6\%, meaning that it is approximately 40\% less effective than the background. Similarly, in Fig.3(e), the curves are very close with AUC about 0.48, indicating that senior agents handling high complexity contacts are as effective as average agents handling normal complexity contacts. In Figure 3(f), we compare the control group as the benchmark and treatment as the target. The AUC is only 0.29, indicating that senior agents are 41.2\% more effective than average agents in handling high complexity contacts. 

The analysis conducted can be expanded to encompass various product lines, as illustrated in Figure 2 in the appendix. The benchmark for this analysis is the background group, while the target group comprises contacts associated with different products or services such as Tablet, Smart Speaker, TV Stick and so on. To ensure the confidentiality of business information, the specific names of the products have been withheld. Through discussions with service agents, we have confirmed that Product-1 consistently receives the highest complexity scores. This is primarily attributed to the fact that most issues related to Product-1 require troubleshooting and cannot be resolved through simple return or refund processes.

\section{Conclusion}
We have developed an intelligent routing mechanism based on a teacher-student learning framework. The teacher model profiles the complexity of a contact based on the customer and agent conversation transcript, which we then use as a data annotator to provide labels to train a student model that predicts complexity using pre-contact features such as customer profile, purchase history, contact history, and device usage statistics. Our experiments have shown that both the teacher and student models are highly aligned with our expectations regarding the properties of complexity and can significantly improve customer experience.

Furthermore, we proposed a novel machine learning metric called Complexity AUC, which evaluates the effectiveness of customer service in handling a particular group of contacts. The metric provides us with a new opportunity to identify bottlenecks in our intelligent routing system and evaluates its effectiveness. By using Complexity AUC, we can gain insights into how the routing business works and make data-driven decisions to further optimize the routing mechanism.

\bibliographystyle{apalike}
\bibliography{citation.bib}

\begin{thebibliography}{}

\bibitem[Balyan et~al., 2020]{difficulty1}
Balyan, R., McCarthy, K.~S., and McNamara, D.~S. (2020).
\newblock Applying natural language processing and hierarchical machine
  learning approaches to text difficulty classification.
\newblock {\em International Journal of Artificial Intelligence in Education
  volume}, 30:337--370.

\bibitem[Balyan et~al., 2021]{difficulty2}
Balyan, R., McCarthy, K.~S., and McNamara, D.~S. (2021).
\newblock Comparing machine learning classification approaches for predicting
  expository text difficulty.
\newblock {\em International Florida Artificial Intelligence Research Society
  Conference}.

\bibitem[Box and Cox, 1964]{power-2}
Box, G. and Cox, D. (1964).
\newblock An analysis of transformations.
\newblock {\em Journal of the Royal Statistical Society B}, 26:211--252.

\bibitem[Chen and Guestrin, 2016]{xgboost}
Chen, T. and Guestrin, C. (2016).
\newblock Xgboost: A scalable tree boosting system.
\newblock {\em ACM SIGKDD International Conference}.

\bibitem[Fawcett, 2006]{auc-3}
Fawcett, T. (2006).
\newblock An introduction to roc analysis.
\newblock {\em Pattern Recognition Letters}, 27(8):861--874.

\bibitem[Gilchrist, 2000]{quantile}
Gilchrist, W. (2000).
\newblock Statistical modelling with quantile functions.
\newblock {\em CRC Press}.

\bibitem[Guo and Berkhahn, 2016]{enity_embedding}
Guo, C. and Berkhahn, F. (2016).
\newblock Entity embeddings of categorical variables.
\newblock {\em arXiv}, page 1604.06737.

\bibitem[Hand and Till, 2001]{auc-2}
Hand, D.~J. and Till, R.~J. (2001).
\newblock A simple generalisation of the area under the roc curve for multiple
  class classification problems.
\newblock {\em Machine Learning}, 45(2):171--186.

\bibitem[Ke et~al., 2017]{LightGBM}
Ke, G., Meng, Q., Finley, T., Wang, T., Chen, W., Ma, W., Ye, Q., and Liu,
  T.-Y. (2017).
\newblock Lightgbm: A highly efficient gradient boosting decision tree.
\newblock {\em Neural Information Processing Systems (NIPS)}.

\bibitem[Kleiman and Page, 2019]{AUC_mu}
Kleiman, R.~S. and Page, D. (2019).
\newblock Auc$_{\mu}$: A performance metric for multi-class machine learning
  models.
\newblock {\em International Conference on Machine Learning (ICML)}, PMLR(97).

\bibitem[Kullback and Leibler, 1951]{kl_divergence}
Kullback, S. and Leibler, R. (1951).
\newblock On information and sufficiency.
\newblock {\em Annals of Mathematical Statistics}, 22:79--86.

\bibitem[McClish, 1989]{auc-1}
McClish, D.~K. (1989).
\newblock Analyzing a portion of the roc curve.
\newblock {\em Med Decis Making}, 9(3):190--195.

\bibitem[Pi et~al., 2023]{universal_model}
Pi, S.-T., Hsieh, C.-P., Liu, Q., and Zhu, Y. (2023).
\newblock Universal model in online customer service.
\newblock {\em WWW '23 Companion: Companion Proceedings of the ACM Web
  Conference 2023}, pages 878--885.

\bibitem[Prokhorenkova et~al., 2018]{catboost}
Prokhorenkova, L., Gusev, G., Vorobev, A., Dorogush, A.~V., and Gulin, A.
  (2018).
\newblock Catboost: unbiased boosting with categorical features.
\newblock {\em d Conference on Neural Information Processing Systems
  (NeurIPS)}.

\bibitem[Rajaraman and Ullman, 2011]{tf-idf}
Rajaraman, A. and Ullman, J.~D. (2011).
\newblock Mining of massive datasets.
\newblock {\em Cambridge University Press}, pages 1--17.

\bibitem[Shannon, 1948]{info_entropy}
Shannon, C.~E. (1948).
\newblock A mathematical theory of communication.
\newblock {\em Bell System Technical Journal}, 27:379--423.

\bibitem[Tishby et~al., 2000]{bottleneck1}
Tishby, N., Pereira, F.~C., and Bialek, W. (2000).
\newblock The information bottleneck method.
\newblock {\em arXiv}, page physics/0004057.

\bibitem[Tishby and Zaslavsky, 2015]{bottleneck2}
Tishby, N. and Zaslavsky, N. (2015).
\newblock Deep learning and the information bottleneck principle.
\newblock {\em arXiv}, page 1503.02406.

\bibitem[Wang, 2006]{difficulty3}
Wang, Y. (2006).
\newblock Automatic recognition of text difficulty from consumers health
  information.
\newblock {\em Computer-Based Medical Systems, 2006. CBMS 2006. 19th IEEE},
  pages 131--136.

\bibitem[Yeo and Johnson, 2000]{power-1}
Yeo, I. and Johnson, R. (2000).
\newblock A new family of power transformations to improve normality or
  symmetry.
\newblock {\em Biometrika}, 87(4):954--959.

\end{thebibliography}

  

\newpage
\renewcommand\thefigure{\thesection.\arabic{figure}}   
\setcounter{figure}{0}
\begin{figure*}
  \centering
  \includegraphics[width=0.8\textwidth]{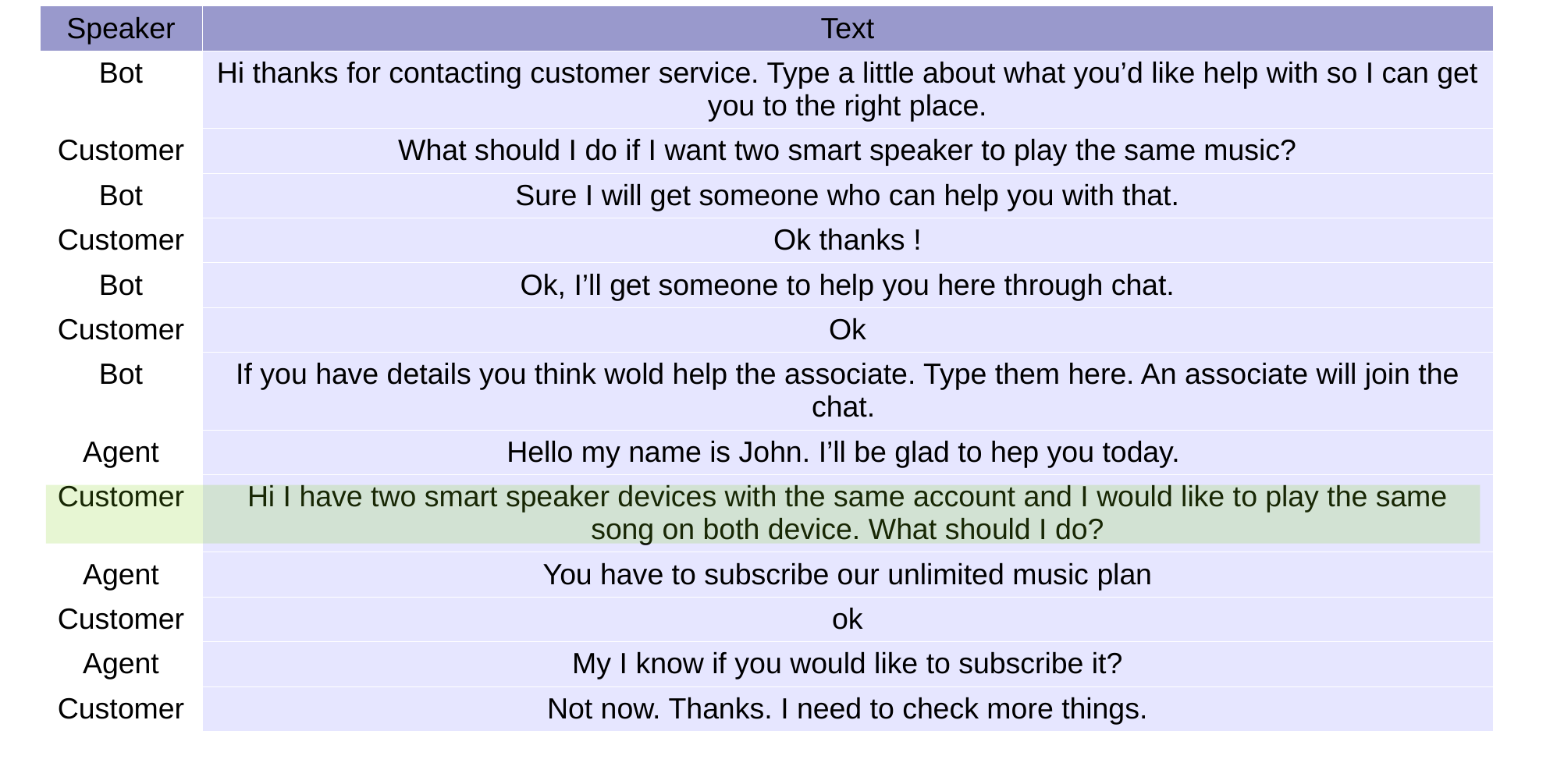}
  \caption{Below is a synthetic example conversation transcript created by a researcher and not sourced from real interactions. The highlighted sentence represents the primary issue raised by the customer. This transcript has been labeled with "Music Related Service" as a unique identifier based on the primary question.}
  \label{example}
\end{figure*}

\begin{figure*}
  \centering
  \includegraphics[width=1.0\textwidth]{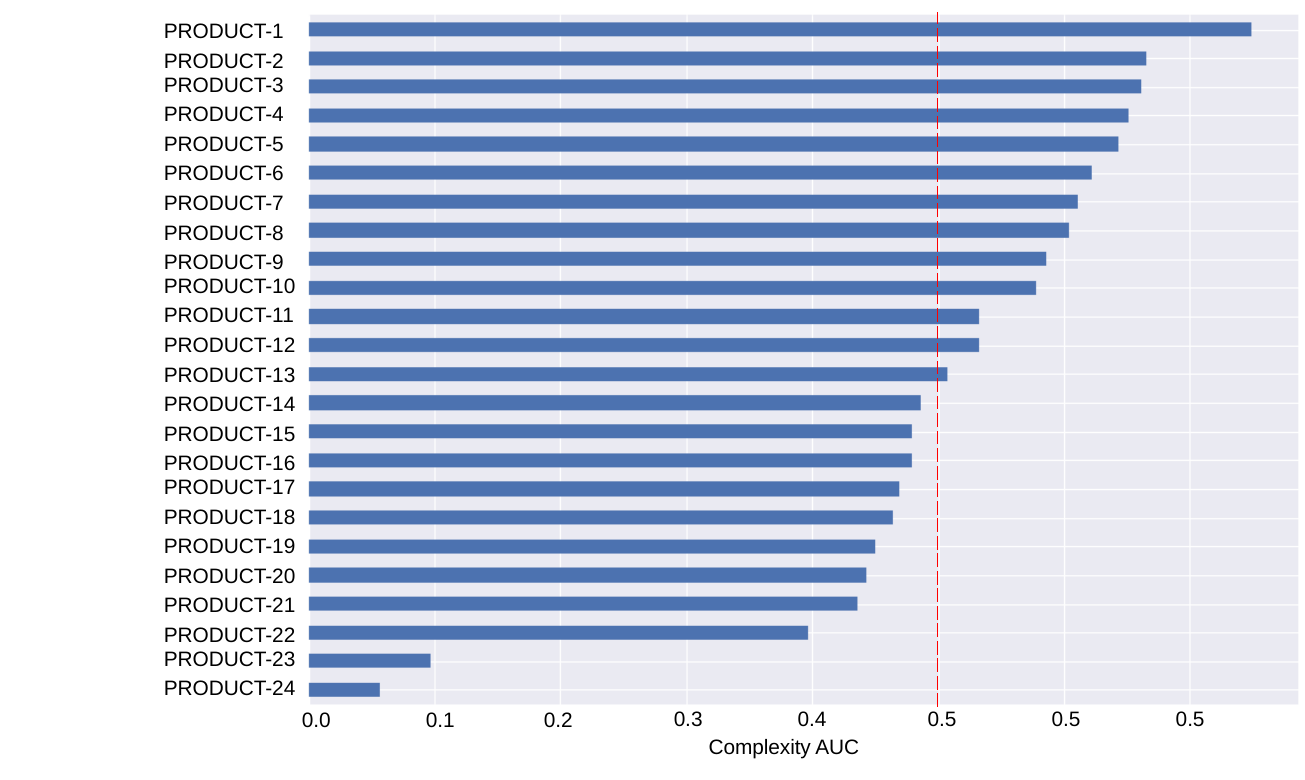}
  \caption{The Complexity AUC is displayed for various product lines or services. The red dashed line represents an AUC of 0.5. Values above the dashed line indicate less effective in serving customers than the average. }\label{skill_cauc}
\end{figure*}
\end{document}